\documentclass[runningheads]{llncs}
\usepackage[T1]{fontenc}
\usepackage{graphicx,verbatim}
\usepackage{amsmath, amssymb}
\usepackage{cite}
\usepackage{multirow}
\usepackage{pifont}
\usepackage{hyperref}
\begin{document}
\title{CoRe-DA: Contrastive Regression for Unsupervised Domain Adaptation in Surgical Skill Assessment}
%

\author{
Dimitrios Anastasiou\inst{1,2} \and
Razvan Caramalau\inst{1} \and
Jialang Xu\inst{1,2} \and
Runlong He\inst{1,2} \and
Freweini Tesfai\inst{4} \and 
Matthew Boal\inst{4} \and 
Nader Francis\inst{4} \and 
Danail Stoyanov\inst{1,3} \and 
Evangelos B. Mazomenos\inst{1,2}}
\authorrunning{Dimitrios Anastasiou et al.}
%
\institute{UCL Hawkes Institute, University College London, UK \and
Dept of Medical Physics \& Biomedical Engineering, University College London, UK \and
Dept of Computer Science, University College London, UK \and
The Griffin Institute, UK
\\
\email{dimitrios.anastasiou.21@ucl.ac.uk}}
\maketitle              
\begin{abstract}
Vision-based surgical skill assessment (SSA) enables objective and scalable evaluation of operative performance. Progress in this field is constrained by the high cost and time demands for manual annotation of quantitative skill scores, as well as the poor generalization of existing regression models to new surgical tasks and environments. Meanwhile, appreciable volumes 
of unlabeled video data are now available, 
motivating the development of unsupervised domain adaptation (UDA) methods for SSA. We introduce the first benchmark for UDA in SSA regression, spanning four datasets across dry-lab and clinical settings as well as open and robotic surgery. We evaluate eight representative models under challenging domain shifts and propose CoRe-DA, a novel contrastive regression–based adaptation framework. Our method learns domain-invariant representations through relative-score supervision and target-domain self-training. Comprehensive experiments across two UDA settings show that CoRe-DA is superior to state-of-the-art methods, achieving Spearman Correlation Coefficients of 0.46 and 0.41 on dry-lab and clinical target datasets, respectively, without using any labeled target data for training. Overall, CoRe-DA enables scalable SSA with reliable cross-domain generalization, where existing methods underperform.
Our code and datasets will be released at \href{https://anonymous.4open.science/r/CoRe-DA-F852}{CoRe-DA}.

\keywords{Surgical Skill Assessment  \and Unsupervised Domain Adaptation \and Contrastive Regression \and Surgical Video Analysis}

\end{abstract}
\section{Introduction}

Automated surgical skill assessment (SSA) has emerged as a promising approach for objective and scalable evaluation of surgical performance \cite{Zia2018}, yet its adoption is constrained by the scarcity of labeled data \cite{Anastasiou2026,Yanik2024}. SSA is typically formulated as a regression task, where annotations are usually produced using standardized frameworks such as the Objective Structured Assessment of Technical Skill (OSATS) \cite{Martin1997}, which rates multiple skill components on 5-point Likert scales. Producing such labels is labor-intensive, as raters (expert surgeons) must repeatedly review videos and coordinate to reach consensus, leading to a substantial annotation cost.
Meanwhile, although current state-of-the-art (SOTA) SSA regression models report strong performance on dry-lab and clinical datasets \cite{Anastasiou2023,Li2022,Ding2023,Liu2021}, they remain largely domain-specific and generalize poorly to new surgical procedures and environments (see Table~\ref{table_main_results}). This makes large-scale deployment challenging under realistic annotation budgets.
In contrast, the volume of unlabeled surgical recordings has been progressively increasing \cite{Wei2025}. This motivates us to address the following question: \textit{Can we learn an SSA regression model that generalizes reliably to a \textbf{target} domain when trained only with labeled \textbf{source} domain data and \textbf{unlabeled} target data? In other words, how far can we go with the data and labels already available?}

\begin{figure*}[t]
    \centering
    \includegraphics[width=1\textwidth]{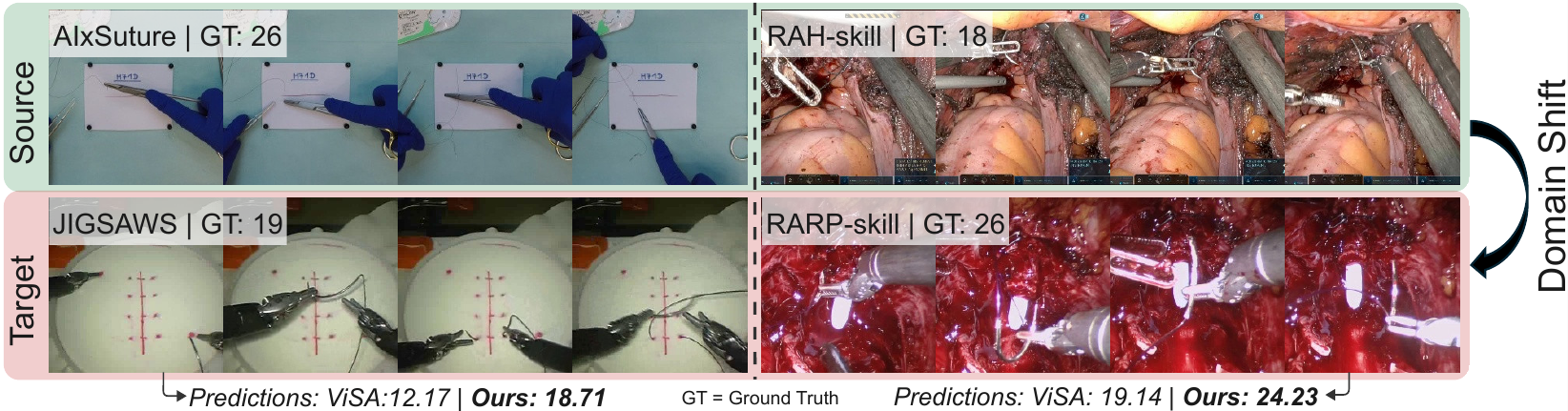}
    \caption{Sample sequences illustrating substantial domain shifts for surgical skill assessment (SSA), with target predictions from ViSA (SOTA) and CoRe-DA (ours).} \label{fig_datasets}
\end{figure*}

Unsupervised Domain Adaptation (UDA) finds direct application in this problem. UDA leverages labeled data from a source
domain and unlabeled data from a target domain to improve model generalization in the target domain \cite{Xu2024}. 
Homologous to our regression task, prior work on UDA includes RSD \cite{Chen2021}, which aligns domains using orthogonal bases of their representation spaces, and DARE-GRAM \cite{Nejjar2023}, which performs adaptation by matching inverse Gram matrices of feature representations. 
Prior work on UDA in SSA is limited \cite{Wang2023,Gong2025} and formulates SSA as binary classification, which fundamentally oversimplifies the task and limits the granularity of feedback that can be provided to surgeons. In contrast, we formulate  SSA prediction as regression (score $\in \mathbb{R}$), enabling fine-grained and more informative assessment of surgical performance. Moreover, \cite{Wang2023} relies on robotic kinematic data, which are typically unavailable, and is evaluated only on simulated environments, while \cite{Gong2025} is exclusive to ophthalmology.

We extend the SOTA by introducing the first formulation of UDA for SSA regression, and establishing a benchmark comprising two UDA settings across four datasets spanning open, robotic dry-lab tasks and clinical robotic surgery (see Fig.~\ref{fig_datasets}), evaluated with eight representative models. These benchmarks present substantial domain shifts due to differences in procedures, acquisition setups, surgical instruments, and visual appearance, making adaptation particularly challenging. Additionally, we propose \textbf{CoRe-DA}, a novel contrastive regression–based UDA framework for SSA which, to the best of our knowledge, is the first to explore contrastive regression as an adaptation strategy. Our formulation is motivated by the observation that standardized SSA tools (\textit{e.g.,} OSATS) assess domain-agnostic performance criteria across procedures and environments, yet regression models struggle to directly estimate performance scores in cross-domain settings. Instead, differences in surgical actions defining skill across videos provide a more reliable signal for learning domain-invariant representations and guiding adaptation. Unlike adversarial \cite{Zhang2019} or feature-alignment approaches \cite{Chen2021,Nejjar2023}, CoRe-DA learns domain-invariant representations by contrasting pairs of videos drawn from both source and target domains. It is trained using absolute and relative score regression losses on source data and regularized through a consistency constraint between the regression heads. In addition, CoRe-DA incorporates self-training on target samples, enabled by pseudo-labels, which is pivotal for driving adaptation and stabilizing regression under severe domain shift. Our main contributions are:

\begin{enumerate}
    \item We formulate UDA for SSA regression and introduce a comprehensive benchmark comprising two challenging UDA settings across four diverse datasets and eight representative baseline models.
    \item We propose CoRe-DA, a novel contrastive regression–based UDA framework for SSA. CoRe-DA represents the first application of contrastive regression as an adaptation strategy in UDA.
    \item Extensive experiments show that CoRe-DA clearly outperforms SOTA methods, achieving improvements in Spearman's Correlation Coefficient (SCC) of $+0.13$ and $+0.26$, and reductions in Mean Absolute Error of 0.32 and 0.49 on dry-lab and clinical robotic surgery target datasets, respectively.
\end{enumerate}

\section{Methods}
\subsection{Preliminaries}
UDA uses labeled data from a source domain and unlabeled data from a target domain to improve model generalization in the target domain. 
We are given a labeled source dataset $\mathcal{D}_S = \{(x_S^i, y_S^i)\}_{i=1}^{N_S}$ and an unlabeled target dataset $\mathcal{D}_T = \{x_T^i\}_{i=1}^{N_T}$, with a common label space. Our goal is to learn a regression model that generalizes to the target domain under distribution shift.

\subsection{Framework (CoRe-DA)}

\noindent \textbf{Overview:} CoRe-DA, shown in Fig.~\ref{fig_framework}, samples batches of triplets $(x_S, x_E, x_T)$ during training, consisting of two (different) labeled source videos, termed \textit{source} and \textit{exemplar}, and an unlabeled \textit{target} video.
A shared encoder $\mathcal{F}$ maps these inputs to feature representations, processed by a relative regressor $\mathcal{R}_{\mathrm{rel}}$ to produce score differences between video pairs (relative score), and an absolute regressor $\mathcal{R}_{\mathrm{abs}}$, to produce individual (absolute) scores. CoRe-DA learns domain-invariant representations via contrastive regression by comparing source–exemplar and target–exemplar pairs. Training uses supervised losses on source predictions together with a self-training strategy on target samples to further drive adaptation.

\noindent \textbf{Clip Sampling and Model Architecture:}
To promote temporal feature diversity, we employ stochastic clip sampling during training as in~\cite{Wang2019,Xu2023}. Given a video of $L$ frames, we divide it into $K$ non-overlapping equal-length segments and randomly sample a clip of $l$ consecutive frames from each, producing a tensor $x \in \mathbb{R}^{(K \cdot l) \times c \times h \times w}$, with $c$, $h$, and $w$ denoting channels, height, and width. We use $x$ to refer to both the full video and its sampled representation.
To align with prior UDA work \cite{Costa2022}, we adopt I3D \cite{Carreira2017} as the feature encoder $\mathcal{F} : \mathbb{R}^{(K \cdot l) \times c \times h \times w} \to \mathbb{R}^{K \times d}$, where $d$ is the feature dimension. The absolute regressor $\mathcal{R}_{\mathrm{abs}} : \mathbb{R}^{K \times d} \to \mathbb{R}$ is implemented via global average pooling (GAP) over the temporal ($K$-clip) dimension, followed by a 3-layer MLP. Following~\cite{Bai2022}, the relative regressor $\mathcal{R}_{\mathrm{rel}}$ operates on concatenated pooled features of video pairs $x_i$ and $x_j$ (\textit{i.e.}, $\mathrm{concat}[\text{GAP}(\mathcal{F}(x_i));\text{GAP}(\mathcal{F}(x_j)]$)), producing a mapping $\mathcal{R}_{\mathrm{rel}} : \mathbb{R}^{K \times 2d} \to \mathbb{R}$. Similarly to $\mathcal{R}_{\mathrm{abs}}$, $\mathcal{R}_{\mathrm{rel}}$ employs a 3-layer MLP.
\newline
\newline
\noindent \textbf{Contrastive Regression-based Domain Adaption:}
For a triplet $(x_S, x_E, x_T)$ we compute absolute and relative score predictions as:
\begin{figure*}[t]
    \centering
    \includegraphics[width=1\textwidth]{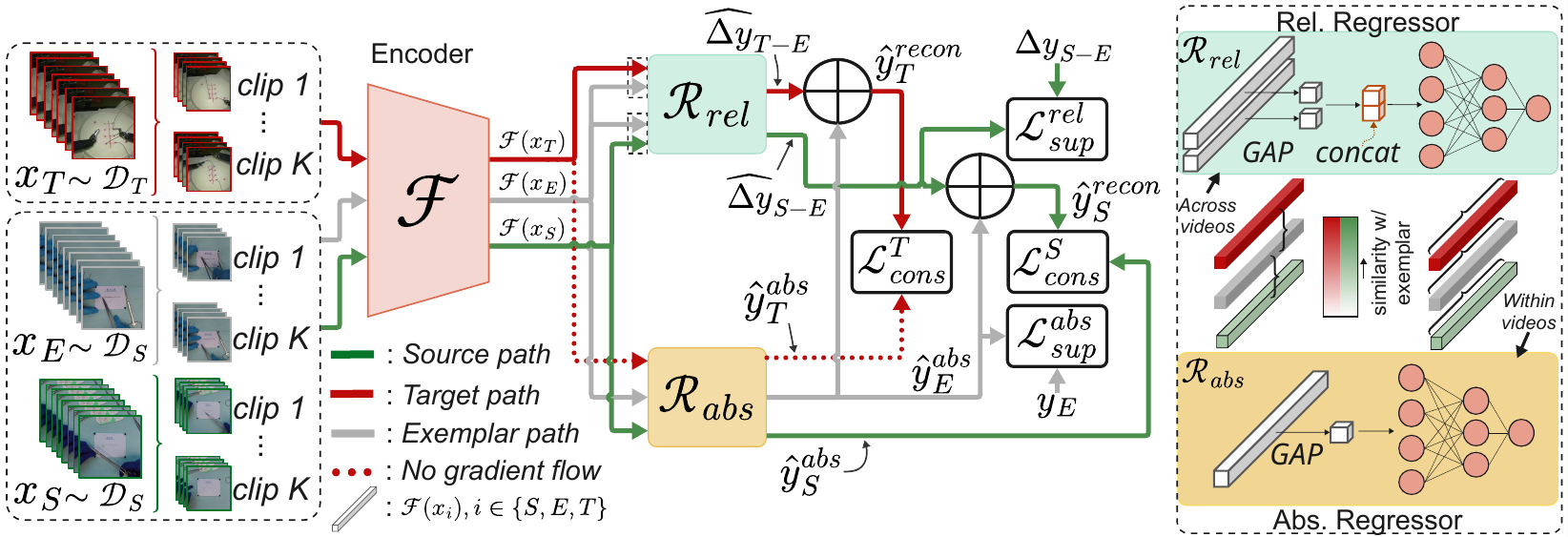}
    \caption{Overview of the proposed CoRe-DA framework. Training uses source–exemplar–target triplets processed by a shared encoder and absolute and relative regression heads. Domain adaptation is driven by contrastive regression between source–exemplar and target–exemplar pairs and pseudo-label self-training.} \label{fig_framework}
\end{figure*}
\begin{align}
\hat{y}_{i}^{\mathrm{abs}} &= \mathcal{R}_{\mathrm{abs}}\!\left(\mathcal{F}(x_i)\right),
\quad i \in \{S,E,T\}, \\
\widehat{\Delta y}_{j-E} &= \mathcal{R}_{\mathrm{rel}}\!\left(\mathcal{F}(x_j), \mathcal{F}(x_E)\right),
\quad j \in \{S,T\}. \label{eq_rel_score}
\end{align}
Considering $\widehat{\Delta y}_{i-j} \triangleq \hat{y}_i - \hat{y}_j$, we \textit{reconstruct} absolute predictions from relative ones as $\hat{y}_{j}^{\mathrm{recon}} = \widehat{\Delta y}_{j-E} + \hat{y}_{E}^{\mathrm{abs}}, j \in \{S,T\}$. 

CoRe-DA is trained on labeled source samples with two supervised losses:  $
\mathcal{L}_{\mathrm{sup}}^{\mathrm{rel}}=
\frac{1}{B}\sum_{i=1}^{B}
(\widehat{\Delta y}_{S-E}^{(i)} - \Delta y_{S-E}^{(i)})^2$ and $\mathcal{L}_{\mathrm{sup}}^{\mathrm{abs}}=
\frac{1}{B}\sum_{i=1}^{B}(\hat{y}_{E}^{\mathrm{abs},(i)} - y_{E}^{(i)})^2$, where $\Delta y_{S-E} = y_S - y_E$, $y_S$ and $y_E$ are the \textit{source} and \textit{exemplar} score labels, and $B$ is the batch size.
$\mathcal{L}_{\mathrm{sup}}^{\mathrm{abs}}$ encourages $\mathcal{F}$ to produce score-discriminative representations~\cite{Samarasinghe2023}, and ensures that the score scale is preserved. Importantly, $\mathcal{L}_{\mathrm{sup}}^{\mathrm{rel}}$ promotes domain-invariant feature learning through contrastive supervision on relative predictions. To avoid prediction drift between the two regressors, we further enforce the following consistency constraint $\mathcal{L}_{\mathrm{cons}}^{S}
=\frac{1}{B}\sum_{i=1}^{B}(\hat{y}_{S}^{recon,(i)} - \hat{y}_{S}^{abs,(i)})^2$.

\begin{figure*}[t]
    \centering
    \includegraphics[width=1\textwidth]{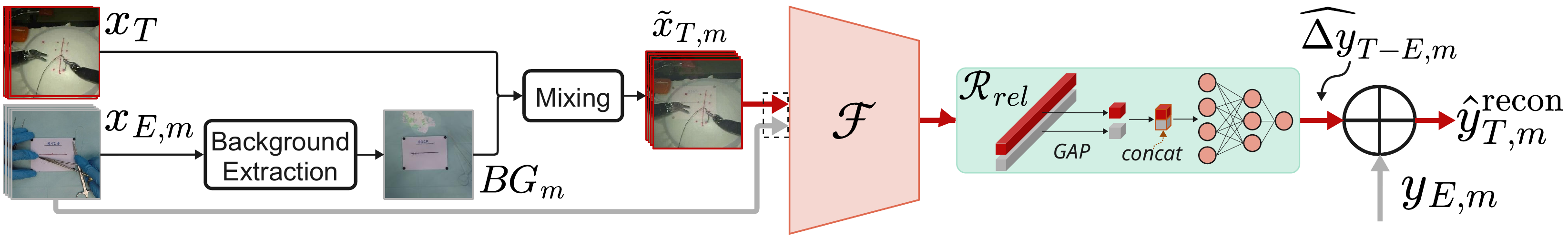}
    \caption{Target-domain testing: Target videos are mixed with exemplar backgrounds, and predictions are reconstructed from relative scores computed across multiple pairs.} \label{fig_inference}
\end{figure*}

Target-domain adaptation is driven by a self-training objective aligning absolute predictions from $\mathcal{R}_{\mathrm{abs}}$ with reconstructed predictions via $\mathcal{R}_{\mathrm{rel}}$. Rather than relying on multiple augmented views of the same input, as in standard consistency-based methods \cite{VanEngelen2020}, our method derives supervision from two complementary prediction pathways (absolute and relative). Inspired by~\cite{Tarvainen2017}, we stop the gradient flow through the absolute prediction branch when computing this loss to prevent model collapse. This process can also be interpreted as $\mathcal{R}_{\mathrm{abs}}$ generating pseudo-labels for target-domain supervision. The resulting target-domain consistency loss is defined as $\mathcal{L}_{\mathrm{cons}}^{T}
=
\frac{1}{B}\sum_{i=1}^{B}(\hat{y}_{T}^{recon,(i)} - \mathrm{stopgrad}(\hat{y}_{T}^{abs,(i)}))^2$.
The total loss is given by $\mathcal{L}_{\mathrm{total}}
=
\alpha (
\mathcal{L}_{\mathrm{sup}}^{\mathrm{rel}} +
\mathcal{L}_{\mathrm{sup}}^{\mathrm{abs}}
)
+
\beta \mathcal{L}_{\mathrm{cons}}^{S}
+
\gamma \mathcal{L}_{\mathrm{cons}}^{T}$,
where $\alpha$, $\beta$, and $\gamma$ are scaling hyperparameters.
\newline
\newline
\noindent \textbf{Target Domain Testing:} As shown in Fig.~\ref{fig_inference}, we employ $\mathcal{R}_{rel}$ using multiple exemplars as in~\cite{Yu2021}. To improve robustness to domain shift, we further apply a test-time transformation in which exemplar backgrounds are mixed with target videos~\cite{Sahoo2021}. Given $M$ exemplar videos $\{x_{E,m}\}_{m=1}^{M}$, sampled uniformly from $\mathcal{D}_S$ and stratified by score label, we generate a single background frame for each using temporal median filtering as in~\cite{Sahoo2021}, resulting in $\{BG_m\}_{m=1}^{M}$. Each background is then mixed with all frames of the target video $x_T$, as $\tilde{x}_{T,m} = (1-\lambda)\, x_T + \lambda\, BG_m$, where $\lambda \in [0,1]$ controls the mixing ratio. For each pair $(\tilde{x}_{T,m}, x_{E,m})$, we produce relative score predictions using Eq.~\ref{eq_rel_score} and reconstruct target scores as $\hat{y}_{T,m}^{\mathrm{recon}}
=
\widehat{\Delta y}_{T-E,m}
+
y_{E,m}$, where $y_{E,m}$ is the exemplar label. The final target prediction is the average over all exemplars: $\bar{\hat{y}}_{T}
=
\frac{1}{M}
\sum_{m=1}^{M}
\hat{y}_{T,m}^{\mathrm{recon}}$.

\section{Experiments}

\noindent \textbf{Benchmark Datasets and UDA Settings:}
We benchmark UDA for SSA on two settings across four datasets: AIxSuture~\cite{Hoffmann2024}, JIGSAWS~\cite{Ahmidi2017}, RARP-skill \cite{He2026}, and RAH-skill \cite{He2026}. AIxSuture is a dry-lab open-surgery dataset of 314 suturing videos annotated with OSATS across 8 components (each scored 1-5). JIGSAWS comprises 103 dry-lab robotic videos of suturing, needle passing, and knot tying, with 6-component OSATS labels. RARP-skill is a 33-video subset of SAR-RARP50~\cite{Psychogyios2023} of dorsal venous complex suturing, annotated with the Modified Global Evaluative Assessment of Robotic Skills (M-GEARS: 6 components, each scored 1-5)~\cite{Boal2024}. RAH-skill contains 37 robotic hysterectomy videos of vaginal cuff closure, each with three suturing tasks, yielding 110 annotated clips with M-GEARS. Based on these datasets, we define two UDA settings under the constraints that $\mathcal{D}_S$ and $\mathcal{D}_T$ share a common label space and that $\mathcal{D}_S$ is sufficiently large for supervised training~\cite{Wilson2020}. The first uses $\mathcal{D}_S=\text{AIxSuture}$ and $\mathcal{D}_T=\text{JIGSAWS}$, keeping their 6 overlapping OSATS components. The second uses $\mathcal{D}_S=\text{RAH-skill}$ and $\mathcal{D}_T=\text{RARP-skill}$, with the same 6-components M-GEARS. The sum of all components is the final score label ($\in[6,30]$). 
\newline
\newline
\noindent \textbf{Evaluation Protocol and Implementation Details:}
Following standard UDA protocols~\cite{Zhang2019,Chen2021}, for each UDA setting, we use all \textbf{labeled} source data and all \textbf{unlabeled} target data for training, and test on all target data. In line with~\cite{Anastasiou2023,Li2022}, we report test performance with  SCC and Mean Absolute Error (MAE).
We initialize I3D with Kinetics weights, and set $d=256$. Both regressors use 3-layer MLPs with output dimensions $256/128/1$. Videos are downsampled to 1\,fps and resized to $224 \times 224$. For training, we sample $l=12$ frames from $K=12$ segments. CoRe-DA is trained with Adam for 150 epochs using learning rates $10^{-5}$ for I3D and $5 \times 10^{-5}$ for $\mathcal{R}_{\mathrm{abs}}$ and $\mathcal{R}_{\mathrm{rel}}$, batch size $16$, and $\alpha{=}\beta{=}\gamma{=}1$. For testing, we sample consecutive non-overlapping clips ($l=12$) to cover the full video. We use $M=10$ exemplars, and set $\lambda=0.25$. CoRe-DA is implemented in PyTorch and trained on an RTX A6000 (48\,GB).

\section{Results}

\noindent \textbf{Comparison with the SOTA:}
\begin{table}[t]
\centering
\caption{UDA results on our two settings (labeled source $+$ unlabeled target), averaged over four runs with different random seeds. Top two results are in \textbf{bold} and \underline{underlined}.}
\label{table_main_results}
{\small
\begin{tabular}{l|c|cc|cc}
\hline
\multirow{2}{*}{Method}      & \multirow{2}{*}{\begin{tabular}[c]{@{}c@{}}Target \\ Adaptat.\end{tabular}} & \multicolumn{2}{c|}{AIxSuture$\to$JIGSAWS}                 & \multicolumn{2}{c}{RAH-skill$\to$RARP-skill}               \\ \cline{3-6} 
                                      &                                                                               & \multicolumn{1}{c|}{SCC ($\uparrow$)} & MAE ($\downarrow$) & \multicolumn{1}{c|}{SCC ($\uparrow$)} & MAE ($\downarrow$) \\ \hline
Source-Only                           & \ding{55}                                                                            & \multicolumn{1}{c|}{0.20 ($\pm$0.15)} & 4.84 ($\pm$0.36)   & \multicolumn{1}{c|}{0.03 ($\pm$0.04)} & 2.89 ($\pm$0.32)   \\ \hline
Contra-Sformer \cite{Anastasiou2023} & \ding{55}                                                                            & \multicolumn{1}{c|}{0.14 ($\pm$0.11)} & 7.45 ($\pm$1.14)   & \multicolumn{1}{c|}{0.03 ($\pm$0.06)} & 2.66 ($\pm$0.39)   \\
ViSA \cite{Li2022}                   & \ding{55}                                                                            & \multicolumn{1}{c|}{0.07 ($\pm$0.26)} & 6.12 ($\pm$0.70)   & \multicolumn{1}{c|}{0.06 ($\pm$0.06)} & 3.10 ($\pm$0.66)   \\ \hline
MDD \cite{Zhang2019}                 & \checkmark                                                                           & \multicolumn{1}{c|}{0.12 ($\pm$0.05)} & 5.02 ($\pm$0.16)   & \multicolumn{1}{c|}{\underline{0.15} ($\pm$0.01)} & 3.02 ($\pm$0.48)   \\
DARE-GRAM \cite{Nejjar2023}          & \checkmark                                                                           & \multicolumn{1}{c|}{0.29 ($\pm$0.04)} & 4.80 ($\pm$0.09)   & \multicolumn{1}{c|}{0.14 ($\pm$0.05)} & 2.54 ($\pm$0.05)   \\
RSD \cite{Chen2021}                  & \checkmark                                                                           & \multicolumn{1}{c|}{0.30 ($\pm$0.02)} & \underline{4.61} ($\pm$0.10)   & \multicolumn{1}{c|}{0.07 ($\pm$0.05)} & \underline{2.42} ($\pm$0.36)   \\
CO$^2$A \cite{Costa2022}                & \checkmark                                                                           & \multicolumn{1}{c|}{\underline{0.33} ($\pm$0.08)} & 5.04 ($\pm$0.72)   & \multicolumn{1}{c|}{0.13 ($\pm$0.06)} & 3.60 ($\pm$0.28)   \\ \hline
\textbf{CoRe-DA (ours)}               & \checkmark                                                                           & \multicolumn{1}{c|}{\textbf{0.46} ($\pm$0.03)} & \textbf{4.29} ($\pm$0.08)   & \multicolumn{1}{c|}{\textbf{0.41} ($\pm$0.05)} & \textbf{1.93} ($\pm$0.07)   \\ \hline
\end{tabular}%
}
\end{table}
\begin{figure*}[t]
    \centering
    \includegraphics[width=1\textwidth]{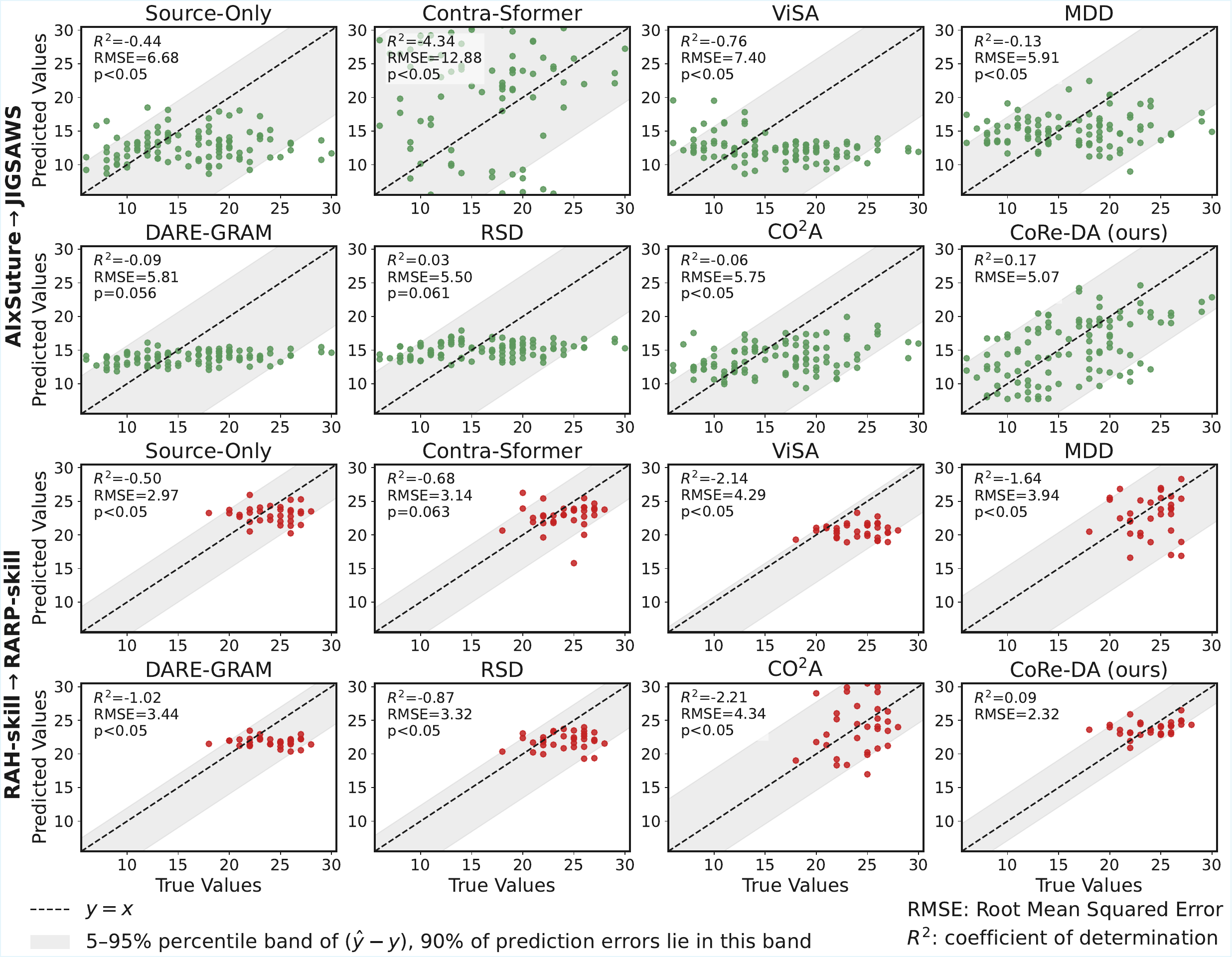}
    \caption{Scatter plots of predicted versus true scores. Paired t-tests show that CoRe-DA outperforms all competing methods with significant or marginally significant improvements (see $p$-values), further supported by higher $R^2$ and lower RMSE.} \label{fig_scatter_plots}
\end{figure*}
We benchmark UDA performance for SSA and compare Core-DA against established UDA baselines: MDD~\cite{Zhang2019}, DARE-GRAM~\cite{Nejjar2023}, RSD~\cite{Chen2021}, and CO$^2$A~\cite{Costa2022}. We also include domain-specific SOTA SSA models, ViSA~\cite{Li2022} and Contra-Sformer~\cite{Anastasiou2023}, and a Source-Only baseline (I3D$+$regressor as in ~\cite{Costa2022}), all trained on labeled source data and evaluated on the target domain without adaptation.
All methods are implemented from released code and adapted to our task; \cite{Gong2025} is omitted, due to code unavailability. For fairness, we use I3D in all UDA methods, while keeping the original architectures for ViSA and Contra-Sformer.
As shown in Table~\ref{table_main_results}, most UDA methods outperform the Source-Only baseline, confirming the benefit of leveraging unlabeled target data. 
CoRe-DA yields the best performance, improving SCC by $+0.13$ and MAE by -$0.32$ in AIxSuture $\rightarrow$JIGSAWS, and boosting SCC by $+0.26$ with a MAE reduction of -$0.49$ in RAH-skill$\rightarrow$RARP-skill, compared to second-best baselines.
Contra-Sformer and ViSA generalize poorly to target domains, reflecting the limitations of domain-specific SSA models under domain shift. CoRe-DA is the only method achieving consistently, with low standard deviation, strong performance across both metrics and settings. 
Fig.~\ref{fig_scatter_plots} shows scatter plots from a single run. AIxSuture$\rightarrow$JIGSAWS proves particularly challenging due to the large domain gap and wider score range, with most methods collapsing predictions toward the mean. 
In contrast, CoRe-DA covers a substantially wider score range, preserves the score ordering, and has smaller error fluctuations, as reflected by a narrower percentile band. Similar trends are observed for RAH-skill$\rightarrow$RARP-skill, where CoRe-DA produces more monotonic predictions with consistently lower errors. We further evaluate a 10-shot semi-supervised UDA setting on AIxSuture$\rightarrow$JIGSAWS, where labels from 10 target videos are available during training and are incorporated with a supervised $\ell_2$ loss. These samples are excluded from testing. In Table~\ref{table_few_shot_results}, CoRe-DA outperforms all other methods, demonstrating its practical value when only a small number of target domain annotations is available.
\begin{table}[t]
\centering
\begin{minipage}[t]{0.48\columnwidth}
\centering
\caption{10-shot Semi-Supervised UDA results on AIxSuture$\to$JIGSAWS, averaged over five labeled-target subsets.}
\label{table_few_shot_results}
\footnotesize

\begin{tabular}{l|c|c}
\hline
\multirow{2}{*}{Method}                      & \multirow{2}{*}{SCC ($\uparrow$)} & \multirow{2}{*}{MAE ($\downarrow$)} \\
                                                      &                                                             &                                                               \\ \hline
Sour.$+$Targ.                                           & 0.39 ($\pm$0.08)                                              & 4.50 ($\pm$0.42)                                                \\ \hline
Contra-S. \cite{Anastasiou2023} & 0.34 ($\pm$0.19)                                              & 5.20 ($\pm$0.84)                                                \\
ViSA \cite{Li2022}                   & 0.32 ($\pm$0.03)                                              & 4.48 ($\pm$0.27)                                                \\ \hline
MDD \cite{Zhang2019}                 & 0.36 ($\pm$0.05)                                              & 4.44 ($\pm$0.30)                                                \\
DARE. \cite{Nejjar2023}          & 0.40 ($\pm$0.07)                                              & \underline{4.14} ($\pm$0.15)                                                \\
RSD \cite{Chen2021}                  & 0.42 ($\pm$0.07)                                              & 4.40 ($\pm$0.17)                                                \\
CO$^2$A \cite{Costa2022}                & \underline{0.46} ($\pm$0.05)                                              & 4.31 ($\pm$0.32)                                                \\ \hline
\textbf{CoRe-DA}                               & \textbf{0.52} ($\pm$0.04)                                              & \textbf{3.83} ($\pm$0.21)                                                \\ \hline
\end{tabular}

\end{minipage}
\hfill
\begin{minipage}[t]{0.50\columnwidth}
\centering
\caption{Ablation studies on AIxSuture$\to$JIGSAWS. Results correspond to a single run using the same random seed.}
\label{table_ablations}
\resizebox{!}{0.28\textwidth}{
\begin{tabular}{cccccc|c|c}
\hline
\multirow{2}{*}{$\mathcal{L}_{sup}^{rel}$} & \multirow{2}{*}{$\mathcal{L}_{sup}^{abs}$} & \multirow{2}{*}{$\mathcal{L}_{cons}^S$} & \multirow{2}{*}{$\mathcal{L}_{cons}^T$} & \multirow{2}{*}{$\operatorname{\substack{\mathrm{stop}\\\mathrm{grad}}}$} & \multirow{2}{*}{$BG_m$} & \multirow{2}{*}{SCC} & \multirow{2}{*}{MAE} \\
                         &                          &                         &                         &                           &                      &                      &                      \\ \hline
\ding{55}                   & \checkmark                & \checkmark               & \checkmark               & \checkmark                 & \checkmark            & 0.38                 & 4.89                 \\
\checkmark                & \ding{55}                   & \checkmark               & \checkmark               & \checkmark                 & \checkmark            & 0.17                 & 5.64                 \\
\checkmark                & \checkmark                & \ding{55}                  & \checkmark               & \checkmark                 & \checkmark            & 0.44                 & 4.57                 \\
\checkmark                & \checkmark                & \checkmark               & \ding{55}                  & \ding{55}                    & \checkmark            & 0.29                 & 4.75                 \\
\checkmark                & \checkmark                & \checkmark               & \checkmark               & \ding{55}                    & \checkmark            & 0.33                 & 4.66                 \\
\checkmark                & \checkmark                & \checkmark               & \checkmark               & \checkmark                 & \ding{55}               & 0.43                 & 4.57                 \\ \hline
\checkmark                & \checkmark                & \checkmark               & \checkmark               & \checkmark                 & \checkmark            & \textbf{0.47}        & \textbf{4.41}        \\ \hline
\end{tabular}%
}

\end{minipage}

\end{table}
\newline
\newline
\noindent \textbf{Ablation Studies: } We evaluate the impact of key CoRe-DA components in Table~\ref{table_ablations}. Without $\mathcal{L}_{\mathrm{sup}}^{\mathrm{rel}}$ performance drops substantially, showing that relative score supervision is essential for learning domain-invariant representations. Removing $\mathcal{L}_{\mathrm{sup}}^{\mathrm{abs}}$ further degrades performance,  
as $\mathcal{R}_{\mathrm{abs}}$ no longer learns the score scale reliably, affecting the quality of target pseudo-labels. $\mathcal{L}_{\mathrm{cons}}^{S}$ has a small but consistent effect as the two regressors may gradually converge to similar solutions. Without $\mathcal{L}_{\mathrm{cons}}^{T}$ target adaptation is effectively disabled, causing a substantial performance drop which highlights the importance of our self-training design for successful domain adaptation. Stopping gradient flow through the absolute branch is essential, as performance otherwise degrades likely due to model collapse.
Test-time background mixing improves adaptation by partially aligning low-level visual statistics across domains. Lastly, increasing the number of exemplars from 2 to 12 monotonically improves SCC (0.43–0.47) and MAE (4.79–4.40), plateauing at 12.

\section{Conclusions}

We present the first benchmark for UDA in SSA regression, consisting of two UDA settings and four datasets across dry-lab and clinical environments. We introduce CoRe-DA, a novel contrastive regression–based framework that learns domain-invariant representations via contrastive supervision and target self-training.
CoRe-DA outperforms SOTA methods, achieving SCC of $0.46$ and $0.41$ and MAE of $4.29$ and $1.93$ on JIGSAWS and RARP-skill, respectively, without using any target labels. Our results demonstrate the potential of our method to enable scalable, video-based SSA in both dry-lab and clinical settings. Future work will explore incorporating uncertainty estimation to improve adaptation by filtering unreliable target pseudo-labels.

\bibliographystyle{splncs04}
\bibliography{refs}

\end{document}